\documentclass[letterpaper, 10 pt, journal, twoside]{IEEEtran}

\usepackage{times}
\usepackage{array}
\usepackage{comment}
\usepackage{csquotes}

\usepackage[numbers,sort&compress]{natbib}
\usepackage{multicol}
\usepackage[bookmarks=true]{hyperref}

\usepackage{amsmath}
\usepackage{amssymb}
\usepackage{cancel}
\usepackage{wasysym}
\usepackage{xfrac}
\usepackage{mathrsfs}
\usepackage{euscript}
\usepackage{bbm}
\usepackage{rotating}
\usepackage{makecell}
\usepackage{xcolor}

\DeclareMathAlphabet{\mymathbb}{U}{bbold}{m}{n}

\usepackage{amsthm}

\usepackage{xcolor}
\usepackage{balance}
\usepackage[normalem]{ulem}

\usepackage[font=footnotesize]{caption}
\usepackage{eso-pic}
\usepackage{textpos}
\usepackage{fancyhdr}

\usepackage{graphicx}
\usepackage{caption}
\usepackage{subcaption}
\usepackage{epstopdf}
\usepackage{float}
\usepackage{tikz}
\usepackage{wrapfig}
\usepackage{xcolor}
\newlength{\sepwid}
\usepackage[maxfloats=256]{morefloats}
\usepackage{url}

\usepackage{lipsum}
\usepackage{cuted}

\usepackage{multirow}

\newcommand{\real}{\mathbb{R}}

\newcommand{\wrench}{{\mathcal F}}

\newcommand{\twist}{{\mathcal V}}
\newcommand{\mass}{{\mathfrak{m}}}
\newcommand{\grav}{{\mathfrak{g}}}
\newcommand{\inertia}{{\mathcal{I}}}
\newcommand{\frames}{\operatorname{s}}
\newcommand{\framec}{\operatorname{c}}
\newcommand{\frameb}{\operatorname{b}}

\begin{document}
\bstctlcite{IEEEexample:BSTcontrol}

\title{Cooperative Payload Estimation by a Team of Mocobots}

\author{
Haoxuan Zhang$^{1}$, C. Lin Liu$^{1}$,  Matthew L. Elwin$^{1}$,  Randy A. Freeman$^{2}$, and Kevin M. Lynch$^{3}$

\thanks{Manuscript received: February 6, 2025; Revised May 19, 2025; Accepted July 24, 2025. This paper was recommended for publication by Editor M. Ani Hsieh upon evaluation of the Associate Editor and Reviewers’ comments. This work was partially supported by NSF under Grant CMMI-2024774.}

\thanks{$^{1}$Haoxuan Zhang, C. Lin Liu, and Matthew L. Elwin are with the Department of Mechanical Engineering, Center for Robotics and Biosystems, Northwestern University, Evanston, IL 60208 USA (e-mail: {\tt\footnotesize haoxuan.zhang@u.northwestern.edu}, {\tt\footnotesize lin.liu@u.northwestern.edu}, {\tt\footnotesize elwin@northwestern.edu)}}%

\thanks{$^{2}$Randy A. Freeman is with the Center for Robotics and Biosystems, Department of Electrical and Computer Engineering, Northwestern Institute on Complex Systems, Northwestern University, Evanston, IL 60208 USA (e-mail: {\tt\footnotesize freeman@northwestern.edu)}}%

\thanks{$^{3}$Kevin M. Lynch is with the Department of Mechanical Engineering, Center for Robotics and Biosystems, Northwestern Institute on Complex Systems, Northwestern University, Evanston, IL 60208 USA (e-mail: {\tt\footnotesize kmlynch@northwestern.edu)}}%



} 

\markboth{IEEE Robotics and Automation Letters. Preprint Version. Accepted July, 2025}
{Zhang \MakeLowercase{\textit{et al.}}: Cooperative Payload Estimation by a Team of Mocobots}

\AtBeginShipoutFirst{%
  \AddToShipoutPicture*{%
    \put(0,0){%
      \makebox[\paperwidth][c]{
        \hspace{1pt}
        \raisebox{2em}[0pt][0pt]{%
          \fbox{%
            \parbox{%
              \dimexpr
                2\columnwidth + \columnsep   
              - 2\fboxsep                    
              - 2\fboxrule                   
              + 10pt                         
            \relax}{%
              \raggedright\footnotesize
              © 2025 IEEE. Personal use of this material is permitted. Permission from IEEE must be obtained for all other uses, in any current or future media, including reprinting/republishing this material for advertising or promotional purposes, creating new collective works, for resale or redistribution to servers or lists, or reuse of any copyrighted component of this work in other works.
            }
          }
        }
      }
    }
  }
}

\maketitle

\begin{abstract}
For high-performance autonomous manipulation of a payload by a mobile manipulator team, or for collaborative manipulation with the human, robots should be able to discover where other robots are attached to the payload, as well as the payload's mass and inertial properties. In this paper, we describe a method for the robots to autonomously discover this information. The robots cooperatively manipulate the payload, and the twist, twist derivative, and wrench data at their grasp frames are used to estimate the transformation matrices between the grasp frames, the location of the payload's center of mass, and the payload's inertia matrix. The method is validated experimentally with a team of three mobile cobots, or mocobots.

\end{abstract}


\section{Introduction} 

\begin{figure} 
\centering
$\vcenter{\hbox{\subfloat{\includegraphics[width=0.44\textwidth]{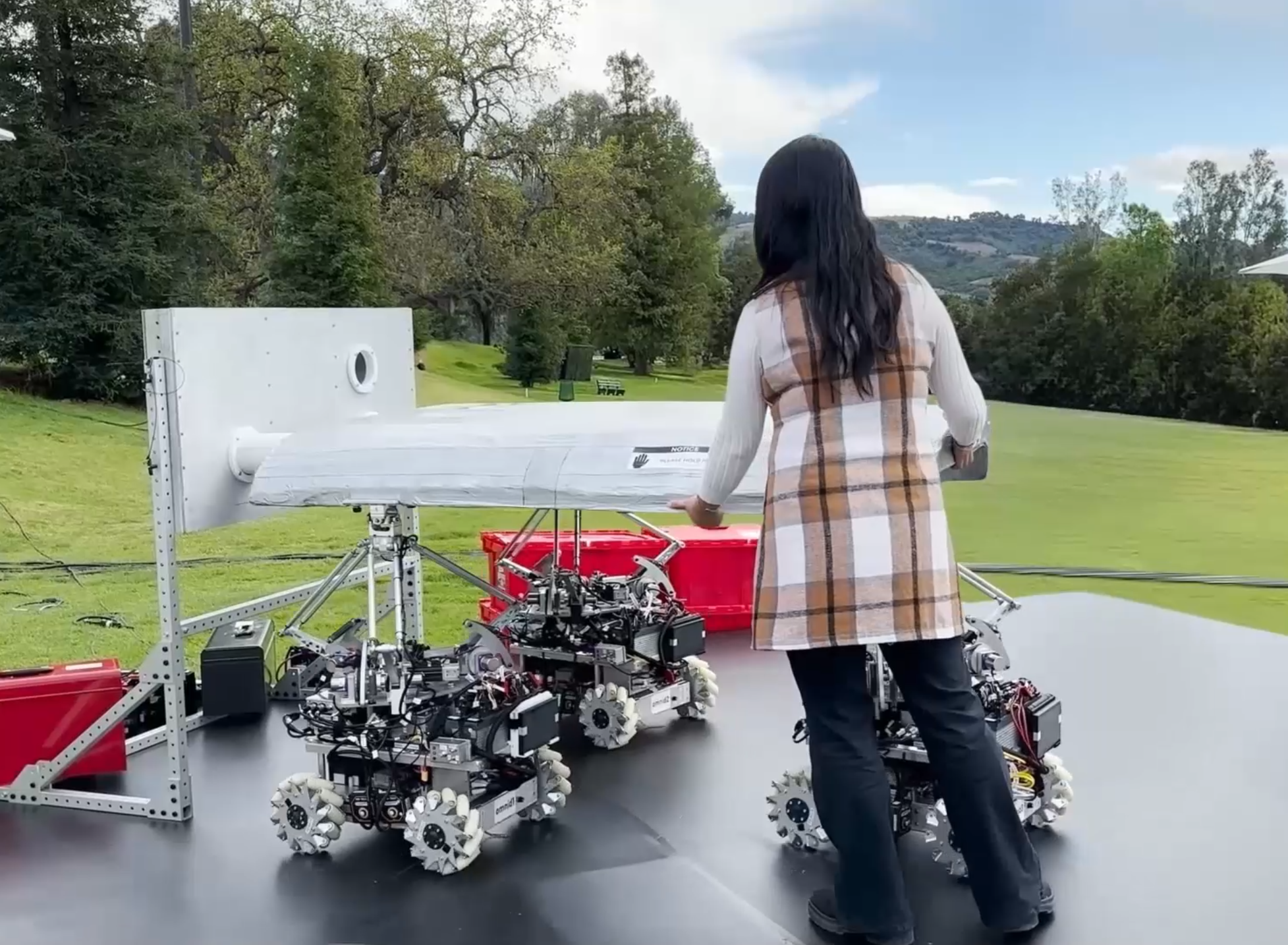}}}}$
\caption{Three Omnid mocobots collaborate safely and intuitively with a human operator on a simulated plane wing assembly task.}
\label{fig:demo}
\end{figure}

\IEEEPARstart{C}{onsider} a scenario in manufacturing, logistics, or construction where a large, substantially rigid payload must be manipulated in all six degrees of freedom (DoF), perhaps for an assembly or loading task. Multiple distributed contacts with the payload are required to respect the wrench limits of any single manipulator and to minimize stress concentrations for heavy or fragile payloads. 
If the manipulation task is not one that is easily automated (e.g., it is not a repetitive task performed in a structured environment), then one or more human operators can physically collaborate with a team of mobile cobots, or mocobots~\cite{elwin2022human}.  The partnership combines the mocobots' physical strength with human perception and adaptability  (Figure~\ref{fig:demo}). 

First, a human guides the mobile manipulators to grasp the common payload. Once the grasps are established, to provide optimal model-based assistance to the human, or for high-performance autonomous manipulation, the robots should be able to discover where the other robots are attached to the payload, as well as the payload's mass and inertial properties. 

In this paper, we describe a method for the robots to autonomously discover this information. The robots cooperatively manipulate the payload, and the twist, twist derivative, and wrench data at their grasp frames are used to estimate the transformation matrices between the grasp frames, the location of the payload's center of mass, and the payload's inertia matrix. The method is validated experimentally with a team of three mocobots.

\subsection{Related Work}
A ``mocobot'' is a mobile variant of a ``cobot,'' robots designed for physical collaboration with humans, originally introduced in~\cite{colgate1996cobots}. Mocobots enhance cobots with mobility for diverse tasks. An example is the Omnid mocobot, designed for human-multirobot collaborative manipulation~\cite{elwin2022human}.

Cooperative robot manipulation involves multiple robots working together to manipulate objects, with a variety of applications in manufacturing, construction, and hazardous environments~\cite{caccavale2016cooperative, cherubini2016collaborative, werfel2014designing, trevelyan2016robotics}. Accurate knowledge of a payload's inertial properties enables planning for dynamic manipulation and high-performance control~\cite{mavrakis2020estimation, mavrakis2016analysis, mavrakis2017safe}. Various cooperative control architectures, including centralized and decentralized controllers, have been proposed for manipulation and transportation under the assumption of known inertial properties~\cite{farivarnejad2022multirobot,verginis2020energy, petitti2016decentralized}. This assumption motivates estimation of such properties for unknown payloads.

A recent survey~\cite{mavrakis2020estimation} provides a comprehensive review of estimation of payload inertial properties, divided into three main categories: purely visual methods, exploratory methods, and methods where the payload is rigidly attached to or grasped by a robot manipulator. The purely visual method relies only on static images or video of the object. For example, the inertial parameters of an object can be estimated visually from the size and shape of the object under assumptions on its density properties~\cite{mirtich1996fast}. Most visual methods require prior knowledge or assumptions, however, such as knowledge of the total mass of the object or an assumption of uniform density.  Exploratory methods include pushing, poking, or tilting an object (e.g.,~\cite{Lynch93estimation,habibi2015distributed}), but motions are often limited to a plane and only a subset of inertial properties can be estimated.  
When a payload is firmly attached to a robot manipulator, Newton-Euler equations can be used to estimate payload inertial properties from robot motion and wrist force-torque data~\cite{atkeson1986estimation}. The basic idea can be extended to grasps by multiple robots, as in~\cite{franchi2014distributed, franchi2018distributed} for a planar payload and~\cite{marino2018} for a 3D payload. The approach and goals of~\cite{marino2018} are closely related to those of this paper, except~\cite{marino2018} does not consider the orientations of the robots' grasp frames and the approach was tested only in simulation.

\subsection{Contribution} 
This paper presents a methodology enabling a group of mobile manipulators to estimate properties of an unknown rigid-body payload. These properties include the payload's mass, center of mass, and inertia matrix, as well as the transformation matrices between the robots' grasp frames, which are initially unknown. 
We have performed extensive experimental validation with mocobots and their force-controlled manipulators, which are well suited to cooperative manipulation and do not have expensive force-torque sensors at their wrists.  Results from these experiments highlight the method's robustness and practicality.

\section{Cooperative Rigid Payload Estimation Problem Formulation}
A team of $N$ robots grasps the rigid body, defining the coordinate frames $\{1\}\ldots \{N\}$ at the grasp locations. Let $T_{i_{0}i} \in SE(3)$ define the transformation matrix describing frame $\{i\}$, the current location of robot $i$'s grasp, relative to a frame $\{i_0\}$, defined as robot $i$'s ``home'' configuration. The configuration
$T_{i_{0}i}$ can be measured by encoders or other sensors on the robot, but the robot has no exteroceptive sensors to directly sense its location relative to other robots or a common world frame.

To determine the properties of the payload and their relative grasp locations, the robots cooperatively manipulate the payload. For example, each robot may attempt to drive its gripper along a periodic reference trajectory using a soft impedance controller. The combination of the reference trajectories should cause the rigid payload to move in all six degrees of freedom, while the impedance control adapts the actual robot trajectories to ensure safe manipulation forces given the unknown connections of the robots to the rigid payload. 

Each robot takes measurements at its grasp interface synchronously with the other robots during manipulation. For example, robot $i$ measures the configuration $T_{i_0i}$, the twist $\twist_i = (\omega_i,v_i) \in \real^6$ measured in $\{i\}$, its time derivative $\dot{\twist}_i$, and the wrench $\wrench_i = (m_i,f_i) \in \real^6$ measured in $\{i\}$. A complete data point for robot $i$ is defined as the tuple $\mathcal{D}_i = \{T_{i_0i}, \twist_{i},\dot{\twist}_{i}, \wrench_{i}\}$.  The set of data points collected by robot $i$ at all $Q$ timesteps is denoted $\mathcal{D}_{i*} = \{ \mathcal{D}_{iq} \; | \; q = 1 \ldots Q \}$, and the set of data points collected by all robots at timestep $q$ is denoted $\mathcal{D}_{*q} = \{ \mathcal{D}_{iq} \; | \; i = 1 \ldots N \}.$ The complete data set is denoted $\mathcal{D}_{**} = \{ \mathcal{D}_{iq} \; | \; i = 1 \ldots N, q = 1 \ldots Q\}$.

The cooperative rigid payload estimation problem can be formulated as follows: given $\mathcal{D}_{**}$, determine the payload's mass $\mass$; the configuration of a frame $\{\framec\}$ at the payload's center of mass relative to each robot's grasp frame ($T_{1c}, \ldots, T_{Nc}$) such that the frame $\{\framec\}$ is aligned with principal axes of inertia of the payload; and the $3 \times 3$ positive-definite inertia matrix $\mathcal{I}$ of the payload in the frame $\{\framec\}$. This information also implies the configuration of each robot's grasp relative to the others, 
\[
T_{ij} = \begin{bmatrix}
    R_{ij} &  p_{ij } \\
    0 & 1
\end{bmatrix} \in SE(3), \; i, j \in \{1, \ldots, N\}.
\]

Payload estimation requires that the data set $\mathcal{D}_{**}$ be sufficiently rich, e.g., the manipulation must cause payload rotations that make estimation of the grasp kinematics and inertial properties well posed. Grasp twist and acceleration data $\twist_i$ and $\dot{\twist}_i$ may be obtained by using filtered encoder data, IMUs, accelerometers, or a combination, and the wrench $\wrench_i$ may be obtained by end-effector force-torque sensors, force/torque sensors at individual robot joints, or other means.

\section{Estimation of Payload Properties}

\subsection{Sequential Estimation}

In principle, all data $\mathcal{D}_{**}$ could be used in a single optimization to simultaneously calculate the configurations of the robot grasps relative to each other, the mass and location of the center of mass of the payload, and the inertia of the payload. In this paper, we adopt a sequenced approach, where we first estimate the grasp kinematics using only twist measurements; then estimate the mass and center of mass using the results of the kinematics estimates and wrench measurements when the payload is held stationary; and finally estimate the inertial properties using the results of the previous estimates as well as twist, acceleration, and wrench data during manipulation. This approach (a) allows us to use simple least-squares estimation and (b) requires only the data needed for the particular parameters being estimated; for example, wrench and acceleration data are not needed to estimate the robots' relative grasp frame configurations. 

\subsection{Grasp Kinematics}
\label{ssec:kinematics}
\begin{figure}
\centering
\includegraphics{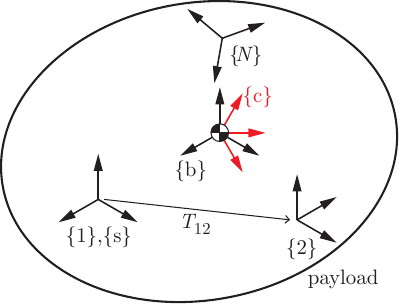}
\caption{The $N$ robots grasp the payload at frames $\{1\}, \ldots, \{N\}$, and one frame (e.g., $\{1\}$) is chosen as a standard reference frame $\{\text{s}\}$. The frame $\{\text{b}\}$ is located at the center of mass and aligned with $\{\text{s}\}$, and $\{\text{c}\}$ is located at the center of mass and aligned with the principal axes of inertia. The transformation matrix $T_{ij}$ locates frame $\{j\}$ relative to $\{i\}$.}
\label{fig:general frame}
\end{figure}

When robots $i$ and $j$ grasp a common rigid body, the twists $\twist_i$ and $\twist_j$ are related by $\twist_i = [\text{Ad}_{T_{ij}}] \twist_j$, where $[\text{Ad}_{T_{ij}}] \in \real^{6 \times 6}$ is the adjoint representation of the transformation matrix $T_{ij}$ describing the configuration of the frame $\{j\}$ relative to $\{i\}$~\cite{lynch2017}. Frames are illustrated in Figure~\ref{fig:general frame}. We write the complete set of twist measurements as  $\twist_{i*},\twist_{j*} \in\mathbb{R}^{6 \times Q}$, i.e., each individual twist measurement $\twist_i$ forms a column of the matrix $\twist_{i*}$. Then
\begin{align}
   \twist_{i*} &= [ \text{Ad}_{T_{ij}} ] \twist_{j*},
\end{align}
or, in expanded form, 
\begin{align}
    \begin{bmatrix}
        \omega_{i*} \\ v_{i*}
    \end{bmatrix}
    & =
    \begin{bmatrix}
        R_{ij} & 0 \\
        [ p_{ij} ] R_{ij} & R_{ij}
    \end{bmatrix}
    \begin{bmatrix}
        \omega_{j*} \\
        v_{j*}
    \end{bmatrix},
\label{eqn:twist}
\end{align}
where $[p_{ij}] \in so(3)$ is the skew-symmetric representation of $p_{ij} \in \real^3$.
This equation is the basis for estimating the rotation matrix $R_{ij}$ and the position vector $p_{ij}$ between the two robots.

\subsubsection{Rotation Matrix}
Let $\omega_{i*}, \omega_{j*} \in \mathbb{R}^{3 \times Q}$ be the angular velocities of the twists measured at each grasp. The relationship between the two sets is
\begin{equation}
    \omega_{i*} = R_{ij} \omega_{j*} \label{eq:rotmat}
.\end{equation}

The problem is to estimate a rotation matrix $\hat{R}_{ij}$ that best fits the data in Equation~\eqref{eq:rotmat} by solving the following least-squares problem based on the Frobenius norm:
\begin{equation}
    \hat{R}_{ij} = \underset{R_{ij} \in SO(3)}{\operatorname{argmin}}  \|R_{ij} \omega_{j*} - \omega_{i*}\| _F ^2.
    \label{eqn:rotation}
\end{equation}
This problem is known as Wahba's problem, which has been solved by the Kabsch-Umeyama algorithm~\cite{kabsch1976,kabsch1978,umeyama1991}.
Defining the matrix $X = \omega_{j*} \omega_{i*}^\intercal$, utilizing SVD decomposition to get $X = U \Sigma V^\intercal$, and defining $S = \operatorname{diag}(1,1,\operatorname{det}(VU^\intercal))$, the solution to the optimization problem is 
\begin{equation}
    \hat{R}_{ij} = VSU^\intercal.
    \label{eqn:rotation2}
\end{equation}
At least three pairs of angular velocities are required to uniquely determine the rotation matrix $R_{ij}$. We assume the measurement data are sufficiently rich to ensure that $X$ is full rank.

\subsubsection{Position Vector}
Let $v_{i*}, v_{j*} \in \mathbb{R}^{3 \times Q}$ be the linear velocities of the twists measured at each grasp. By Equation~\eqref{eqn:twist}, the linear velocities satisfy
\begin{equation}
    v_{i*} = [p_{ij}] R_{ij} \omega_{j*} + R_{ij} v_{j*}.
    \label{eq:linear}
\end{equation}
Due to the skew-symmetric property $[\omega]p=-[p]\omega$, Equation~\eqref{eq:linear} can be rearranged to
\begin{equation}
    v_{i*} = - [ R_{ij} \omega_{j*} ] p_{ij} + R_{ij} v_{j*}.
\end{equation}
Plugging in the estimate $\hat{R}_{ij}$, the estimate $\hat{p}_{ij}$ is found by solving the least-squares problem
\begin{equation}
\hat{p}_{ij} = \underset{p_{ij}}{\operatorname{argmin}} \;  \|
    [ \hat{R}_{ij} \omega_{j*} ] p_{ij} - (\hat{R}_{ij} v_{j*} - v_{i*})\|_2,
    \label{eq:p-solution}
\end{equation}
provided $[\hat{R}_{ij}\omega_{j*}]$ is full rank.

The solution $\hat{T}_{ij} = (\hat{R}_{ij},\hat{p}_{ij})$, via Equations~\eqref{eqn:rotation} and \eqref{eq:p-solution}, depends only on pairwise robot data. To obtain a complete and consistent representation of the robots' relative grasp frames, $N-1$ such solutions are needed, e.g., $\hat{T}_{12}, \hat{T}_{13}, \ldots, \hat{T}_{1N}$. Then the configuration of any frame $\{i\}$ relative to another frame $\{j\}$ may be calculated as $\hat{T}_{ij} = \hat{T}^{-1}_{1i}\hat{T}_{1j}$. While this method permits efficient linear least-squares computation, it does not simultaneously take into account all combinations of robots' twist data nor loop-closure constraints among three or more robots, e.g., $T_{ij}T_{jk}T_{ki} = I$. Such constraints allow squeezing more information out of the collected data at the cost of greater computational complexity and nonlinear optimization.

We employed an iterative gradient-based nonlinear optimization to incorporate all the data and loop constraints to further refine the $\hat{T}_{ij}$ estimates from the original linear least-squares solutions. The optimization minimizes the weighted sum of squared errors from data from all pairwise combinations of grasping frames and all closed loops. Our implementation uses the scipy.optimize library and the Broyden-Fletcher-Goldfarb-Shanno method for estimating gradients~\cite{nocedal2006large}, but other nonlinear optimization methods incorporating combinatorial loop-closure constraints could also be employed~\cite{thrun2005probabilistic, lu1997globally}.

\subsection{Mass and Center of Mass (CoM)}
\label{ssec:com}

The mass and center of mass are estimated using the grasp kinematics solution and a set of static wrench measurements.
A common reference frame $\{\frames\}$ is chosen from among the grasp frames $\{\{1\}, \ldots , \{N\}\}$; for example, $\{\frames\}$ may be chosen to be coincident with $\{1\}$ (Figure~\ref{fig:general frame}).
We also define a frame $\{\frames_0\}$ coincident with $\{\frames\}$ but oriented such that its $\hat{z}$-axis is opposite the gravity vector $\grav \in \real^3$. The center of mass of the payload is located at the origin of a to-be-estimated frame $\{\framec\}$.

With local wrenches $\wrench_i = (m_i,f_i)$ from each robot, the force and moment static equilibrium conditions are

\begin{equation}
\mass \grav = - \sum\limits_{i=1}^{N} R_{s_0i} f_i
\label{eq:mass}
\end{equation}

\begin{equation}
[p_{sc}]  ({R_{ss_0}} \mass \grav) =- \sum\limits_{i=1}^{N} \Big([p_{si}] (R_{si}f_i) + R_{si} m_i \Big),
\label{eq:com}
\end{equation}where $\mass$ is the mass of the payload, $f_i$ is the measured force component of $\wrench_i$, and $p_{sc}$ and $p_{si}$ represent the origin of $\{\framec\}$ and $\{i\}$ in $\{\frames\}$ coordinates, respectively. 

The mass $\mass$ can be estimated using Equation~\eqref{eq:mass} and one or more static measurements by the robots, while $p_{sc}$ can be estimated using Equation~\eqref{eq:com} and two or more static measurements, provided the orientations of the payload during measurement differ by rotation about an axis not aligned with the gravity vector $\grav$.

Since the robots rigidly grasp the payload, $p_{sc}$, $p_{si}$, and $R_{si}$ are constant, and only $p_{sc}$ remains unknown. Given $q = 1 \ldots Q$ measurements by the robots, we define the scalars 
\begin{equation}
A_q = [0,0,1] \grav, \quad
b_q = - [0,0,1] \left(\sum_{i=1}^{N} R_{s_0i}^q f_i^q\right) 
\end{equation}
and the vectors

\begin{equation}
    A = \begin{bmatrix}
        A_1 \\
        A_2 \\
        \vdots \\
        A_Q
    \end{bmatrix} \in \mathbb{R}^{Q \times 1}, \quad
    b = \begin{bmatrix}
        b_1 \\
        b_2 \\
        \vdots \\
        b_Q
    \end{bmatrix} \in \mathbb{R}^{Q \times 1}.
\end{equation}
Plugging this data into the $\hat{z}$-component of Equation~\eqref{eq:mass}, we get
\begin{equation}
A\mass = b,
\end{equation}
which can be solved for the mass estimate $\hat{\mass}$ using least squares.

Similarly, the data can be plugged into Equation~\eqref{eq:com} to estimate the center-of-mass location $p_{sc} \in \real^3$: 
\begin{equation}
A_q = [{R_{ss_0}^q}  \hat{\mass} \grav] \in \real^{3 \times 3},
\end{equation}

\begin{equation}
b_q = \sum_{i=1}^{N} \Big( [p_{si}] (R_{si}^q f_i^q ) + R_{si}^q m_i^q \Big) \in \mathbb{R}^{3 \times 1},
\end{equation}
\begin{equation}
    A = \begin{bmatrix}
        A_1 \\
        A_2 \\
        \vdots \\
        A_Q
    \end{bmatrix} \in \mathbb{R}^{3Q \times 3}, \quad
    b = \begin{bmatrix}
        b_1 \\
        b_2 \\
        \vdots \\
        b_Q
    \end{bmatrix} \in \mathbb{R}^{3Q \times 1},
\end{equation}
\begin{equation}
Ap_{sc} = b,
\end{equation}
which can be solved for the estimate $\hat{p}_{sc} \in \real^3$ using least squares.

\subsection{Inertia Matrix}
\label{ssec:inertia}

We define a frame $\{\frameb\}$, shown in Figure~\ref{fig:general frame}, at the center of mass $\hat{p}_{sc}$ and aligned with $\{\frames\}$. After estimating the inertia matrix in $\{\frameb\}$, we identify the principal axes of inertia in $\{\frameb\}$, and then define a final center-of-mass frame $\{\framec\}$ with axes aligned with the principal axes of inertia. 

The Newton-Euler dynamics of a rotating rigid body are 
\begin{equation}
    m_{b} = \mathcal{I}_b \alpha_{b} +[\omega_{b}] \mathcal{I}_b \omega_{b},
    \label{eqn:inertia1}
\end{equation}
where $m_b$ is the moment acting on the body, $\mathcal{I}_b \in \mathbb{R}^{3 \times 3}$ is the positive-definite inertia matrix of the payload, $\omega_b$ is the angular velocity, and $\alpha_b$ is the angular acceleration, all expressed in the frame $\{\frameb\}$. For $N$ robots supporting the payload 
with local wrenches $\wrench_i = (m_i,f_i)$, the moment can be expressed as 

\begin{equation}
    m_{b} = \sum_{i=1}^{N} \Big([p_{bi}] (R_{bi} f_{i}) + R_{bi} m_i \Big).
    \label{eqn:inertia2}
\end{equation}

Equations~\eqref{eqn:inertia1} and ~\eqref{eqn:inertia2} can be reorganized to solve for $\inertia_b$ using linear least squares. 
Since $\inertia_b$ is symmetric, it contains six unique elements: 
$\inertia_{xx}$, $\inertia_{xy}$, $\inertia_{xz}$, $\inertia_{yy}$, $\inertia_{yz}$, and $\inertia_{zz}$. We define
\begin{equation}
    \inertia_{\operatorname{reg}} = [\inertia_{xx}, \inertia_{xy}, \inertia_{xz}, \inertia_{yy}, \inertia_{yz}, \inertia_{zz}]^\intercal .
\end{equation}

At timestep $q$, we construct the matrices $A_q, B_q \in \real^{3 \times 6}$ (based on the angular acceleration $\alpha_b = (\alpha_x, \alpha_y, \alpha_z)$ and angular velocity $\omega_b$ measured in $\{\frames\}$) and the vector $y_q$ based on force measurements at each of the $N$ robots:
\begin{equation}
    A_q = 
    \begin{bmatrix}
        \alpha_{x} & \alpha_{y} & \alpha_{z} & 0 & 0 & 0 \\
        0 & \alpha_{x} & 0 & \alpha_{y} & \alpha_{z} & 0 \\
        0 & 0 & \alpha_{x} & 0 & \alpha_{y} & \alpha_{z} 
    \end{bmatrix},
\end{equation}
    \begin{align}  
        B_q = 
        &\left[
        \begin{array}{ccc}
        0 & -\omega_{x}\omega_{z} & \omega_{x}\omega_{y} \\
        \omega_{x}\omega_{z} & \omega_{y}\omega_{z} & \omega_{z}^2 - \omega_{x}^2 \\
        -\omega_{x}\omega_{y} & -\omega_{y}^2 + \omega_{x}^2 & -\omega_{y}\omega_{z}
        \end{array}
        \right. \\[10pt]
        & \quad\quad\quad \left.
        \begin{array}{ccc}
        -\omega_{y}\omega_{z} & -\omega_{z}^2 + \omega_{y}^2 & \omega_{y}\omega_{z} \\
        0 & -\omega_{x}\omega_{y} & -\omega_{x}\omega_{z} \\
        \omega_{x}\omega_{y} & \omega_{x}\omega_{z} & 0
        \end{array}
        \right]_, \nonumber
    \end{align}

\begin{equation}
    y_q = \sum_{i=1}^{N} \Big([\hat{p}_{bi}](\hat{R}_{bi}f_{iq}) + \hat{R}_{bi}m_{iq}\Big).
\end{equation}

Combining Equations~\eqref{eqn:inertia1} and~\eqref{eqn:inertia2}, the dynamics can be written 
\begin{equation}
    (A_q + B_q) \inertia_{\operatorname{reg}} = y_q. 
\end{equation}
For $Q$ measurements, we define
\begin{equation}
    X = \begin{bmatrix}
        A_1 + B_1 \\
        A_2 + B_2 \\
        \vdots \\
        A_Q + B_Q
    \end{bmatrix} \in \mathbb{R}^{3Q \times 6}, \quad
    y = \begin{bmatrix}
        y_1 \\
        y_2 \\
        \vdots \\
        y_Q
    \end{bmatrix} \in \mathbb{R}^{3Q \times 1},
\end{equation}
yielding 
\begin{equation}
    X \inertia_{\operatorname{reg}} = y,
\end{equation}
which $\hat{\inertia}_{\operatorname{reg}}$ solves in a least-squares sense. The elements of $\hat{\inertia}_{\operatorname{reg}}$ form the entries of the estimated inertia matrix  $\hat{\inertia}_b$.\footnote{If noisy data causes the least-squares estimate to produce an $\hat{\inertia}_b$ with a negative eigenvalue, the inertia estimate can either be discarded or projected to the space of positive semidefinite matrices by performing an eigenvalue decomposition and replacing any negative eigenvalue with a zero eigenvalue.} 

Let $R_{bc} = [r_1, r_2, r_3]$, where $r_i$ is the $i$th eigenvector of $\hat{\inertia}_b$ and $\{\framec\}$ is a frame coincident with $\{\frameb\}$ with axes aligned with the principal axes of inertia. Then the estimated diagonal inertia matrix in $\{\framec\}$ is 
\begin{equation}
    \hat{\inertia}_c = R_{bc}^\intercal \hat{\inertia}_b R_{bc}
.\end{equation}

\section{Experimental Implementation with the Omnid Mocobots}

\begin{figure}
\centering
\vspace{2pt}
$\vcenter{\hbox{\subfloat{\includegraphics[width=0.4\textwidth]{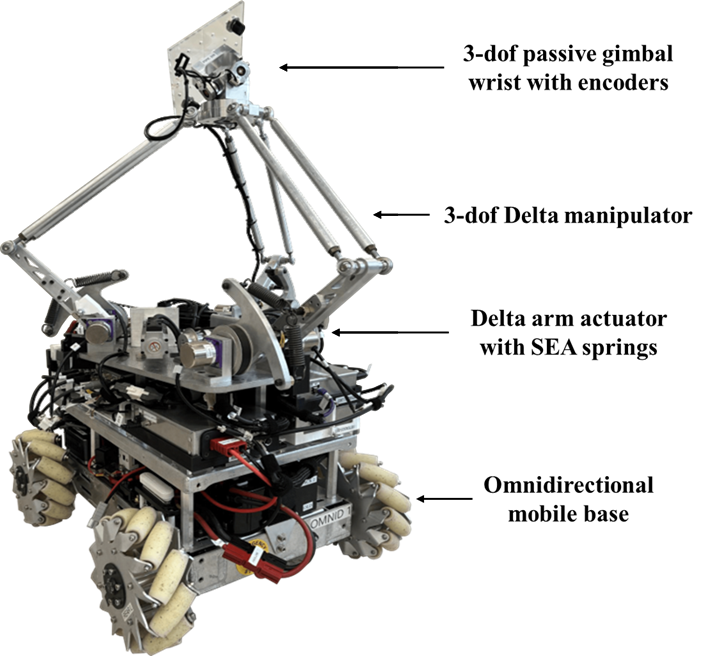}}}}$
\caption{The structure of the Omnid mocobot.}
\label{fig:omnid}
\end{figure}

We performed payload estimation using the Omnid mocobots (Figure \ref{fig:omnid}), which are designed for human-robot collaborative manipulation~\cite{elwin2022human}.  Each Omnid mocobot consists of a mecanum-wheel mobile base, a 3-DoF Delta parallel mechanism driven by series-elastic actuators (SEAs), and a 3-DoF passive gimbal wrist. In this paper, the mobile bases are stationary, and all manipulation is performed by the Delta-plus-gimbal manipulators.

The Delta mechanism and gimbal wrist of each Omnid are equipped with encoders, allowing Omnid $i$ to measure its wrist configuration $T_{i_0i}$. Since the payload is rigidly attached to the gimbal, the wrist configuration and the grasp configuration are equivalent. The twist $\twist_i$ and its derivative $\dot{\twist}_i$ are calculated by filtering encoder data. Wrenches at the wrist take the form $\wrench_i = (m_i,f_i) = (0,f_i)$, where the linear force $f_i$ is calculated based on torques measured at the SEA joints and the moment $m_i$ is zero due to the passive gimbal wrist. Data is collected at a fixed sampling rate of 100~Hz. 

Experimental video is at
\url{https://youtu.be/fAvCyRhzfC4} or 
\url{https://tinyurl.com/mocobot-estimation}.

\begin{figure}
    \centering
    \includegraphics[width=0.45\textwidth]{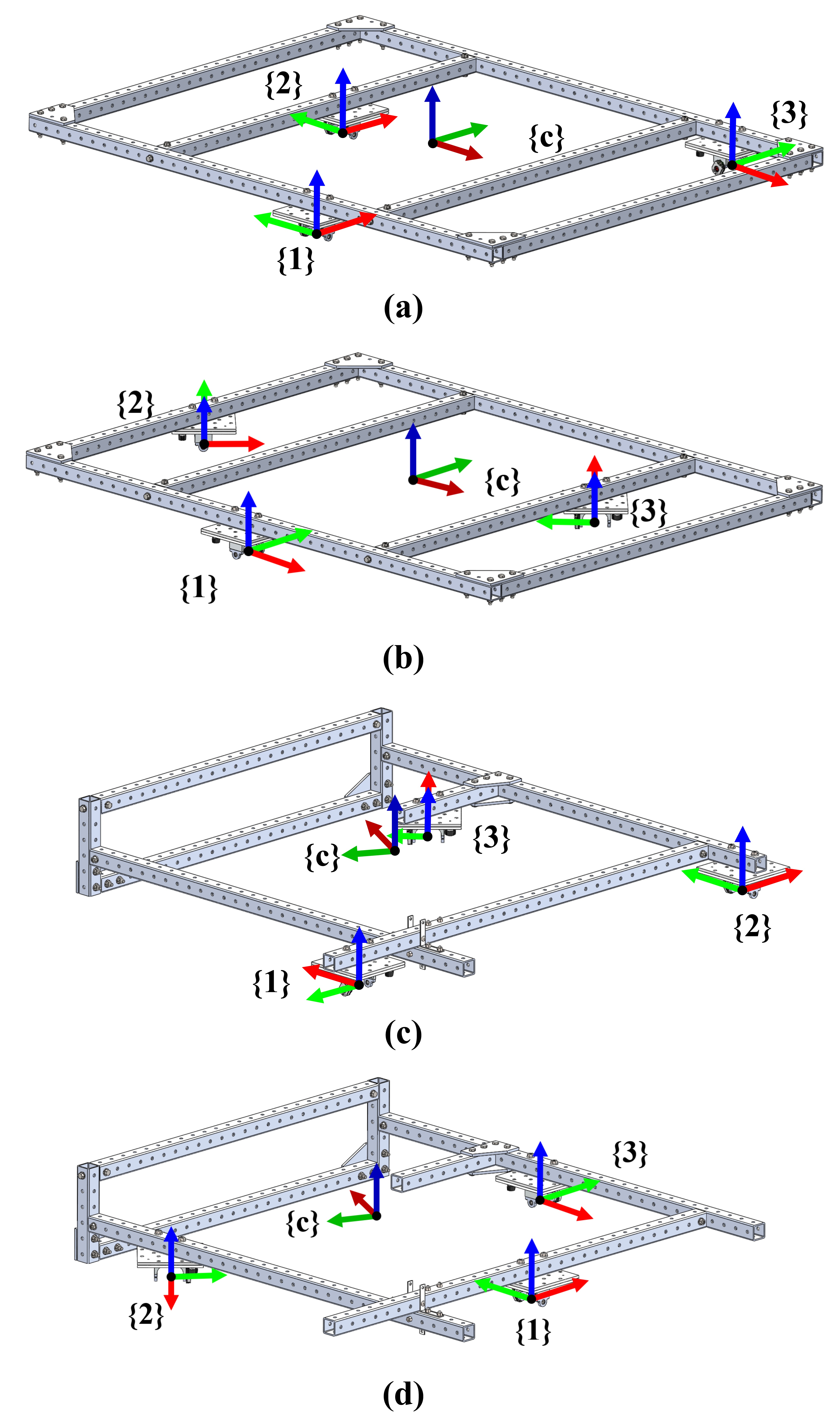}
    \caption{Four payloads with associated grasp frames for experimental validation. Frame $\{1\}$ is chosen as the reference frame $\{\frames\}$. The axes of the frames are visually differentiated by colors, with red corresponding to the $x$-axis, green to the $y$-axis, and blue to the $z$-axis.}
    \label{fig:payload configuration}
\end{figure}

\subsection{Experimental Configuration}
To experimentally validate the methodology, we used four different payloads: two structures constructed from aluminum extrusion, each with two different sets of grasp locations (Figure~\ref{fig:payload configuration}). 

\begin{figure}
    \centering
    \includegraphics[width=0.43\textwidth]{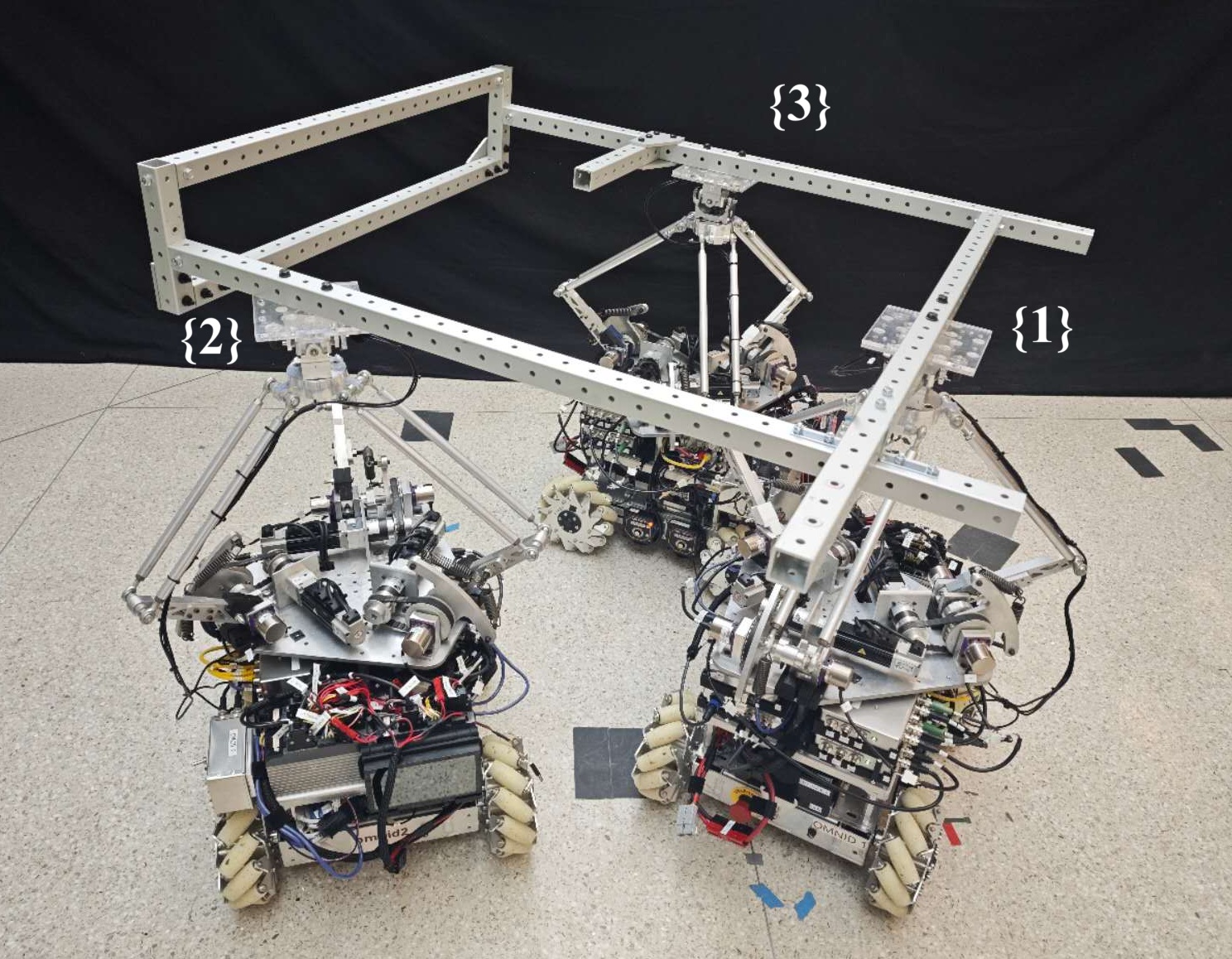}
    \caption{Experimental setup with payload configuration (d).}
    \label{fig:system frame}
\end{figure}

The payloads are rigidly attached to the Omnid gimbal wrists during experiments. The Omnid grasp frames are denoted $\{1\}$, $\{2\}$, and $\{3\}$, and frame $\{1\}$ is selected as the reference frame $\{\frames\}$.   Figure~\ref{fig:system frame} shows the three Omnids carrying payload (d).

The ground truth values of the transformation matrices, mass, center of mass, and inertia matrix for experimental validation were derived from CAD and are summarized in Table~\ref{tab:GT}. Rotation matrices are summarized in the axis-angle (exponential coordinates) representation $\omega \beta \in \real^3$, where $\omega$ is the unit rotation axis and $\beta$ is the angle of rotation, expressed in this paper in degrees.

\begin{table}[t]
    \centering
    \vspace{5pt}
    \renewcommand{\arraystretch}{1.6}
    \setlength{\tabcolsep}{5pt}
    \resizebox{0.48\textwidth}{!}{
    \begin{tabular}{|c|c|c|c|}
        \hline
        \multicolumn{4}{|c|}{\textbf{Rotation Matrix in Axis Angle (deg)}} \\
        \hline
        \textbf{Config.} & $R_{12}$ & $R_{23}$ & $R_{31}$ \\
        \hline
        (a) & (0, 0, 0) & (0, 0, $-$90) & (0, 0, 90) \\
        (b) & (0, 0, 45) & (0, 0, 90) & (0, 0, $-$135) \\
        (c) & (0, 0, $-$90) & (0, 0, 45) & (0, 0, 45) \\
        (d) & (0, 0, $-$135) & (0, 0, 45) & (0, 0, 90) \\
        \hline
        \multicolumn{4}{|c|}{\textbf{Position Vector (m)}} \\
        \hline
        \textbf{Config.} & $p_{12}$ & $p_{23}$ & $p_{31}$ \\
        \hline
        (a) & (0.647, 0.533, 0) & (0.457, $-$0.838, 0) & ($-$0.305, $-$1.105, 0) \\
        (b) & ($-$0.686, 0.572, 0) & (0.916, $-$0.647, 0) & ($-$0.242, 0.835, 0) \\
        (c) & ($-$0.115, $-$1.143, 0) & ($-$0.229, 0.800, 0) & ($-$1.132, 0.162, 0) \\
        (d) & ($-$0.382, 0.800, $-$0.038) & ($-$0.485, 0.862, 0.038) & (0.533, $-$0.571, 0) \\
        \hline
        \multicolumn{4}{|c|}{\textbf{Mass (kg), Center of Mass (m), and Principal Axes (deg)}} \\
        \hline
        \textbf{Config.} & \textbf{$\mass$} & $p_{1c}$ & $R_{1c}$ \\
        \hline
        (a) & 11.672 & (0.537, 0.156, 0.072) & (0.043, 0.085, -84.787) \\
        (b) & 11.672 & (0.034, 0.535, 0.072) & (0.115, $-$0.115, 2.465) \\
        (c) & 10.478 & (0.547, $-$0.670, 0.057) & ($-$1.034, 0.355, $-$24.643) \\
        (d) & 10.478 & (0.089, 0.593, 0.055) & ($-$1.662, $-$0.702, 68.937) \\
        \hline
        \multicolumn{4}{|c|}{\textbf{Inertia Matrix (kg$\cdot$m$^2$)}} \\
        \hline
        \textbf{Config.} & $\inertia_{xx}$ & $\inertia_{yy}$ & $\inertia_{zz}$ \\
        \hline
        (a) & 2.318 & 3.215 & 5.524 \\
        (b) & 2.153 & 3.390 & 5.535 \\
        (c) & 1.824 & 2.438 & 4.208 \\
        (d) & 1.711 & 2.122 & 3.772 \\
        \hline
    \end{tabular}
    }
    \caption{Ground truth values for the four experimental configurations.}
    \label{tab:GT}
\end{table}

\subsection{Data Processing}
\begin{figure}
    \centering
    \includegraphics[width=0.48\textwidth]{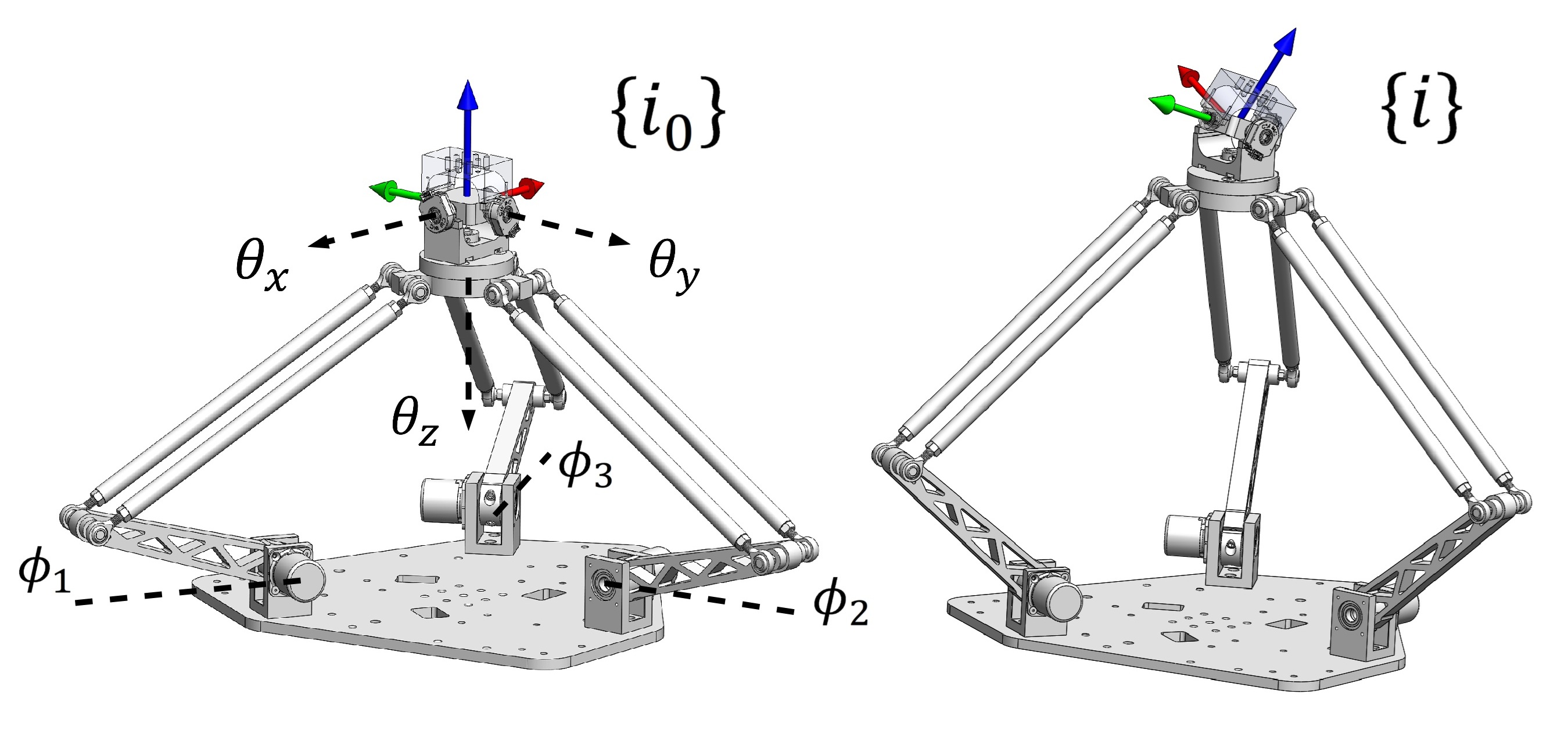}
    \caption{Frame $\{i\}$ of the 6-DoF manipulator of Omnid $i$ is positioned at the center of the gimbal wrist. The $\hat{z}$-axis encoder is installed at the base of the gimbal wrist, which is not visible in the figure. The left image shows the manipulator in its home position.}
    \label{fig:delta frame}
\end{figure}

As shown in Figure~\ref{fig:delta frame}, $\phi = (\phi_1, \phi_2, \phi_3)$ represent the actuated proximal joint angles of the Delta mechanism and $\theta = (\theta_x, \theta_y, \theta_z)$ represent the angles of the gimbal wrist, all measured by encoders. The transformation matrix $T_{i_0 i}$ can be derived through forward kinematics as
\begin{equation}
    T_{i_0 i}(\theta,\phi) = \begin{bmatrix}
                    g(\theta) & h(\phi) \\
                    0 & 1
                \end{bmatrix}
.\end{equation}
The twist $\twist_i = (\omega_i,v_i)$ is calculated as
\begin{equation}
    T_{i_0i}^{-1} \dot{T}_{i_0i}  = [\twist_i] = 
    \begin{bmatrix}
        [\omega_i] & v_i \\
        0 & 0
    \end{bmatrix} \in se(3).
\end{equation}
The wrench at $\{i\}$ is $\wrench_i = (0, f_i)$, where $f_i = (\partial h/\partial \phi)^{-\intercal} \tau_i$ and $\tau_i$ is the Delta's SEA joint torque vector. 

The derivative $\dot{T}_{i_0i}$ and the twist derivative $\dot{\twist}_i$ are calculated using central differencing post-processing of the $100$~Hz estimates of $T_{i_0i}$ and $\twist_i$, respectively. To smooth the data and reduce the effect of amplification of encoder quantization in this numerical differencing, we apply a third-order Butterworth low-pass filter to the encoder data prior to calculating the $T_{i_0i}$ estimates. The same filter is applied to smooth the SEA torque data. To avoid filter edge effects and other transients, the first and last two seconds of each trial's data are discarded.

\subsection{Experimental Results}

\subsubsection{Grasp Kinematics}
To generate sufficient twist information for accurate estimation of relative grasp frame locations, the payload was randomly moved in all six DoF by assigning independent random trajectories to each Omnid manipulator. Each random trajectory consisted of a series of nominally straight-line paths for the gimbal wrist between via points, with the randomized transit times and dwell times at the vias chosen to be between $0.5$ and $0.8$~s. Because these independent random trajectories are not kinematically compatible with rigid grasps on the payload, we took advantage of mechanical compliance and used low motion control gains at each Omnid's force-controlled manipulator to allow trajectory tracking error to keep internal forces on the payload low.

Encoder data from the Omnid manipulators were collected to calculate $\twist_{i*}, i = 1, 2,  3$. The transformations for the individual robot pairs $\hat{T}_{12}$, $\hat{T}_{23}$, and $\hat{T}_{31}$ were first calculated using the least-squares methods of Section~\ref{ssec:kinematics}, and these estimates were then refined using nonlinear optimization incorporating loop-closure constraints. In our experiments, this second optimization step did not change the estimates significantly. Six independent trials were conducted for each payload, and the random trajectories for each trial consisted of $80$ random via points. 

The errors between the estimated transformation matrices and the ground truth transformation matrices (Table~\ref{tab:GT}) are reported in Tables~\ref{tab:rotation} and \ref{tab:position} for the rotation matrices and position vectors, respectively. Rotation matrix errors are expressed as the rotation angle $\beta$ (in degrees) of $[\omega]\beta = \log(R_{ij}^\intercal \hat{R}_{ij})$.

\begin{table}
    \vspace{5pt}
    \centering
    \renewcommand{\arraystretch}{1.6}
    \resizebox{0.35\textwidth}{!}{%
    \begin{tabular}{|c|c|cc|}
        \hline
        \multirow{2}{*}{\textbf{Config.}} 
        & \multirow{2}{*}{\makecell{\textbf{Parameters}\\}}
        & \multicolumn{2}{c|}{\textbf{Absolute Error (deg)}} \\
        \cline{3-4}
        & & \textbf{Mean} & \textbf{Standard Deviation} \\
        \hline
        \multirow{3}{*}{(a)} & $R_{12}$ & 0.94 & 0.21 \\
                             & $R_{23}$ & 0.99 & 0.16 \\
                             & $R_{31}$ & 1.07 & 0.53 \\
        \hline
        \multirow{3}{*}{(b)} & $R_{12}$ & 1.77 & 0.19 \\
                             & $R_{23}$ & 0.90 & 0.42 \\
                             & $R_{31}$ & 1.93 & 0.42 \\
        \hline
        \multirow{3}{*}{(c)} & $R_{12}$ & 0.66 & 0.36 \\
                             & $R_{23}$ & 2.21 & 0.59 \\
                             & $R_{31}$ & 1.67 & 0.70 \\
        \hline
        \multirow{3}{*}{(d)} & $R_{12}$ & 2.55 & 0.27 \\
                             & $R_{23}$ & 3.57 & 0.32 \\
                             & $R_{31}$ & 1.49 & 0.38 \\
        \hline
    \end{tabular}
    }
    \caption{Experimental results for relative grasp frame rotation matrix estimation. Results are represented in absolute degrees of rotation around the axis that rotates the estimated rotation matrix to the actual rotation matrix.}
    \label{tab:rotation}
\end{table}

\begin{table}
    \centering
    \vspace{5pt}
    \renewcommand{\arraystretch}{1.6}
    \resizebox{0.48\textwidth}{!}{%
    \begin{tabular}{|c|c|cc|cc|}
        \hline
        \multirow{2}{*}{\textbf{Config.}}
        & \multirow{2}{*}{\makecell{\textbf{Parameters}\\}}
        & \multicolumn{2}{c|}{\textbf{Absolute Error (m)}} 
        & \multicolumn{2}{c|}{\textbf{Percentage Error}} \\
        \cline{3-6}
        & & \textbf{Mean} & \textbf{Standard Deviation} 
          & \textbf{Mean} & \textbf{Standard Deviation} \\
        \hline
        \multirow{3}{*}{(a)} & $p_{12}$ & 0.007 & 0.003 & 0.9\% & 0.3\% \\
                             & $p_{23}$ & 0.042 & 0.005 & 4.4\% & 0.5\% \\
                             & $p_{31}$ & 0.051 & 0.008 & 4.5\% & 0.7\% \\
        \hline
        \multirow{3}{*}{(b)} & $p_{12}$ & 0.016 & 0.002 & 1.8\% & 0.2\% \\
                             & $p_{23}$ & 0.039 & 0.006 & 3.5\% & 0.6\% \\
                             & $p_{31}$ & 0.039 & 0.002 & 4.5\% & 0.3\% \\
        \hline
        \multirow{3}{*}{(c)} & $p_{12}$ & 0.038 & 0.003 & 3.3\% & 0.2\% \\
                             & $p_{23}$ & 0.046 & 0.004 & 5.6\% & 0.4\% \\
                             & $p_{31}$ & 0.063 & 0.007 & 5.5\% & 0.6\% \\
        \hline
        \multirow{3}{*}{(d)} & $p_{12}$ & 0.019 & 0.002 & 2.1\% & 0.3\% \\
                             & $p_{23}$ & 0.057 & 0.004 & 5.8\% & 0.4\% \\
                             & $p_{31}$ & 0.029 & 0.004 & 3.8\% & 0.5\% \\
        \hline
    \end{tabular}
    }
    \caption{Experimental results for relative grasp frame position estimation. Each 3-vector position error is converted to a scalar using the Euclidean norm, and percentage errors are relative to the Euclidean norm of the ground truth $p_{ij}$.}
    \label{tab:position}
\end{table}

\subsubsection{Mass and Center of Mass}

To estimate the mass and center of mass of the payload, the Omnids were commanded to hold the payload stationary at six distinct orientations. 
Data collection was performed during periods when the measured forces along all axes remained constant (with a tolerance of $0.01$~N) for a continuous duration of six seconds. The results of the least-squares estimates from Section~\ref{ssec:com} are summarized in Table~\ref{tab:CoM}.

\begin{table}[t]
    \centering
    \renewcommand{\arraystretch}{1.6}
    \resizebox{0.38\textwidth}{!}{%
    \begin{tabular}{|c|c|c|c|}
        \hline
        \textbf{Config.} & \textbf{Parameters} & \textbf{Absolute Error} & \textbf{Percentage Error} \\
        \hline
        \multirow{2}{*}{(a)} & $\mass$ (kg) & 0.11 & 0.9\% \\
                             & $p_{1c}$ (m) & 0.021 & 3.7\% \\
        \hline
        \multirow{2}{*}{(b)} & $\mass$ (kg) & 0.07 & 0.6\% \\
                             & $p_{1c}$ (m) & 0.018 & 3.3\% \\
        \hline
        \multirow{2}{*}{(c)} & $\mass$ (kg) & 0.11 & 1.1\% \\
                             & $p_{1c}$ (m) & 0.039 & 4.5\% \\
        \hline
        \multirow{2}{*}{(d)} & $\mass$ (kg) & 0.18 & 1.8\% \\
                             & $p_{1c}$ (m) & 0.023 & 3.8\% \\
        \hline
    \end{tabular}
    }
    \caption{Experimental results for mass and center of mass estimation. The 3-vector position error is converted to a scalar using the Euclidean norm, and the percentage error is relative to the Euclidean norm of the ground truth $p_{1\text{c}}$.}
    \label{tab:CoM}
\end{table}

\subsubsection{Inertia Matrix}
To estimate the inertia matrix, the Omnids were tasked with generating motion to sufficiently excite the payload's rotational dynamics. Two types of trajectories were employed: the random trajectories used previously in the grasp kinematics experiments, and periodic trajectories constructed based on the estimated grasp transformations. For the periodic trajectories, all three Omnid wrists moved along periodic trajectories on the surface of the sphere for which the origins of the three grasp frames lie on a common meridian. In total, six experimental trials for each configuration were conducted: three trials using random trajectories, each consisting of 80 configurations per Omnid, and three trials using controlled periodic trajectories, each lasting one minute. No significant difference in the data quality was observed between the two trajectory types.

Table~\ref{tab:inertia} summarizes the results from the six experimental trials.
Similar to the labeling of axes in the ground truth inertia matrix, the labels $\hat{\inertia}_{xx}$, $\hat{\inertia}_{yy}$, and $\hat{\inertia}_{zz}$ are given to the axes with principal inertia estimates in ascending order.

\begin{table}[t]
    \centering
    \vspace{5pt}
    \renewcommand{\arraystretch}{1.6}
    \resizebox{0.48\textwidth}{!}{%
    \begin{tabular}{|c|c|cc|cc|}
        \hline
        \multirow{2}{*}{\textbf{Config.}}
        & \multirow{2}{*}{\makecell{\textbf{Parameters}\\}}
        & \multicolumn{2}{c|}{\textbf{Absolute Error (\textbf{kg$\cdot$m$^2$})}}
        & \multicolumn{2}{c|}{\textbf{Percentage Error}} \\
        \cline{3-6}
        & & \textbf{Mean} & \textbf{Standard Deviation} 
          & \textbf{Mean} & \textbf{Standard Deviation} \\
        \hline
        \multirow{3}{*}{(a)} & $\inertia_{xx}$ & 0.083 & 0.036 & 3.6\% & 1.6\% \\
                             & $\inertia_{yy}$ & 0.130 & 0.022 & 4.1\% & 0.7\% \\
                             & $\inertia_{zz}$ & 0.102 & 0.166 & 1.8\% & 3.0\% \\
        \hline
        \multirow{3}{*}{(b)} & $\inertia_{xx}$ & 0.067 & 0.065 & 3.1\% & 3.0\% \\
                             & $\inertia_{yy}$ & 0.086 & 0.102 & 2.5\% & 3.0\% \\
                             & $\inertia_{zz}$ & 0.246 & 0.056 & 4.4\% & 1.0\% \\
        \hline
        \multirow{3}{*}{(c)} & $\inertia_{xx}$ & 0.005 & 0.073 & 0.3\% & 4.0\% \\
                             & $\inertia_{yy}$ & 0.039 & 0.120 & 1.6\% & 4.9\% \\
                             & $\inertia_{zz}$ & 0.158 & 0.091 & 3.8\% & 2.2\% \\
        \hline
        \multirow{3}{*}{(d)} & $\inertia_{xx}$ & 0.048 & 0.058 & 2.8\% & 3.4\% \\
                             & $\inertia_{yy}$ & 0.016 & 0.060 & 0.8\% & 2.8\% \\
                             & $\inertia_{zz}$ & 0.005 & 0.165 & 0.1\% & 4.4\% \\
        \hline
    \end{tabular}
    }
    \caption{Experimental results for the inertias of the principal axes of inertia for different payloads. 
    }
    \label{tab:inertia}
\end{table}


\section{Discussion}

The experimental results demonstrate the validity of the approach, producing relative grasp rotation matrix mean errors of less than $4^\circ$ and position vector magnitude mean errors of less than $6\%$, a mass estimate error of less than $2\%$, and a center of mass position error of less than $5\%$.  Inertia matrix estimation, which relies on acceleration estimates from twice-differenced encoder data, exhibits a bit more error, but it is still less than $7\%$ for the principal moments of inertia when accounting for one standard deviation. 

The experimentally-derived kinematic and mass parameters may be directly inserted into the mocobot gravity compensation mode reported in~\cite{elwin2022human}, and the inertial parameters can be used to improve the performance of dynamic model-based cooperative control, eliminating the need for an \textit{a priori} model of the payload and the grasp locations. There are a number of limitations to the method, however.

First, grasp frame velocities and accelerations are obtained by differencing encoder data, including encoders at the proximal joints of the Delta mechanism, distant from the wrist. The intervening mechanism and the differencing process introduce noise to the estimation. To address this, the Omnid sensor suite could be supplemented by IMUs at the gimbals. Independent of the sensing modality, however, the inherent compliance of the manipulators, due to their SEAs, can introduce vibrations at the wrists, which may be exacerbated by a poor choice of wrist trajectories to estimate the payload inertial properties. In practice, our excitation trajectories resulted in relatively smooth motions of the payload, as shown in the video.

Second, manipulation forces at the wrist are estimated based on the Delta's proximal joint SEAs. This measurement assumes zero friction at joints and bearings of the Delta mechanisms, but it eliminates the need for costly and potentially fragile end-effector force-torque sensors. Friction in the Omnid Delta joints is quite low, resulting in good wrist force sensing.

Third, the limited workspace of the manipulators places constraints on the payload trajectories that can be used to determine inertial properties. In particular, the payload trajectories in our experiments were limited to less than $\pm 10^\circ$ rotation about any axis.  Payload trajectories with larger rotations could provide a higher signal-to-noise ratio. 

Finally, computational aspects of the method (e.g., data filtering during post-processing) could be further optimized.

\section{Conclusion}
We present a methodology for cooperative payload estimation using a group of mocobots, focusing on the estimation of the mass, center of mass, and inertia matrix of a rigid body payload, as well as the transformations between the grasp frames of each robot. 
Despite challenges of sensor noise and unmodeled dynamics, the proposed approach demonstrates promising results for payload estimation using Omnid mocobots. 
Future work will focus on extending the methodology to identify the kinematic and inertial properties of articulated payloads, and eventually continuously deformable payloads.

\label{sec:conclusion} 

\bibliographystyle{IEEEtran}
\balance
\small
\Urlmuskip=0mu plus 1mu\relax
\bibliography{Cooperative_Payload_Estimation_by_a_Team_of_Mocobots.bib}

\end{document}